\documentclass[11pt]{article}

\usepackage{booktabs}
\usepackage[final]{acl}
\usepackage{verbatim}
\usepackage{algorithm}
\usepackage{algpseudocode}
\usepackage[compact]{titlesec}
\usepackage{enumitem}
\setlist{nosep, leftmargin=*}
\usepackage{graphicx}
\usepackage{amssymb} 

\usepackage{hyperref} 

\usepackage[table]{xcolor}

\newcommand{\best}[1]{\textbf{#1}}
\newcommand{\second}[1]{\underline{#1}}

\usepackage{multirow}
\usepackage[most]{tcolorbox}
\usepackage{xcolor}

\newtcolorbox{promptbox}[2][]{
  colback=black!3!white,    
  colframe=black!40!white,  
  coltitle=black,           
  title={\textbf{#2}},      
  fonttitle=\small\sffamily, 
  fontupper=\scriptsize\ttfamily, 
  boxrule=0.8pt,            
  arc=2pt,                  
  left=3pt, right=3pt, top=3pt, bottom=3pt, 
  breakable,                
  enhanced,                 
  #1                        
}
\usepackage{amsmath}
\usepackage[english,bidi=default]{babel} 

\title{RISER: Orchestrating Latent Reasoning Skills for Adaptive Activation Steering}

\author{
Wencheng Ye \\ Tongji University \\\And
Xiaoyang Yuan \\ Tongji University \\\And
Yi Bin \\ Tongji University \\\And
Hengyu Jin \\ Tongji University \\\AND
Liang Peng \\ Tencent \\\And
Pengpeng Zeng \\ Tongji University \\\And
Heng Tao Shen \\ Tongji University
}

\begin{document}

\maketitle
\begin{abstract}
Recent work on domain-specific reasoning with large language models (LLMs) often relies on training-intensive approaches that require parameter updates. While activation steering has emerged as a parameter-efficient alternative, existing methods apply static, manual interventions that fail to adapt to the dynamic nature of complex reasoning. To address this limitation, we propose \textbf{RISER} (\textbf{R}outer-based \textbf{I}ntervention for \textbf{S}teerable \textbf{E}nhancement of \textbf{R}easoning), a plug-and-play intervention framework that adaptively steers LLM reasoning in activation space. RISER constructs a library of reusable reasoning vectors and employs a lightweight Router to dynamically compose them for each input. The Router is optimized via reinforcement learning under task-level rewards, activating latent cognitive primitives in an emergent and compositional manner. Across seven diverse benchmarks, RISER yields 3.4--6.5\% average zero-shot accuracy improvements over the base model while surpassing CoT-style reasoning with 2–3× higher token efficiency and robust accuracy gains. Further analysis shows that RISER autonomously combines multiple vectors into interpretable, precise control strategies, pointing toward more controllable and efficient LLM reasoning. Code can be found in: \href{https://github.com/gooogleshanghai/RISER-Orchestrating-Latent-Reasoning-Skills-for-Adaptive-Activation-Steering}{\texttt{RISER}}.

\end{abstract}

\section{Introduction}
\begin{figure}
    \centering
    \includegraphics[width=1\linewidth]{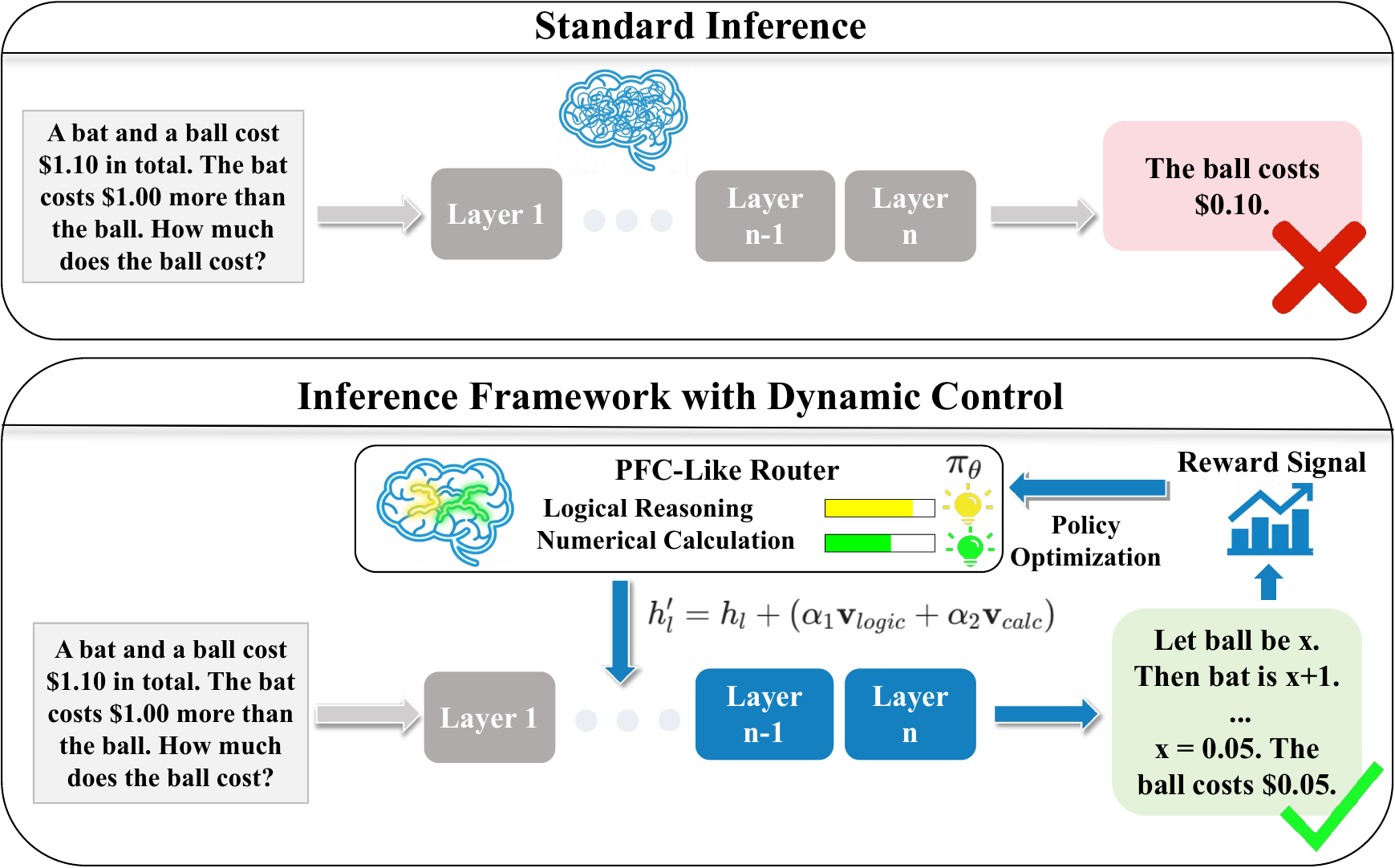}
    \caption{Conceptual comparison between Standard Inference and the RISER framework. RISER (bottom) uses a learned Router to dynamically inject composed vectors, analogous to an explicit executive-control mechanism. 
    }
    \label{fig:concept}
\end{figure}


Large Language Models (LLMs)~\citep{touvron2023llamaopenefficientfoundation,brown2020language} exhibit strong general reasoning abilities~\citep{ahn2024largelanguagemodelsmathematical,wei2022emergentabilitieslargelanguage,shi2024math}, yet they often perform inconsistently on specialized downstream tasks requiring domain knowledge, disciplined reasoning patterns, and the coordinated use of multiple cognitive skills~\citep{wang2025evaluationllmsmathematicalproblem,li202512surveyreasoning,10.1145/3641289}. In practice, we seek to strengthen such reasoning without expensive retraining or relying solely on indirect prompting strategies, motivating mechanisms that can directly modulate the model’s internal computation during inference.


Existing approaches face fundamental limitations. Training-based methods such as Supervised Fine-Tuning (SFT) and Reinforcement Learning (RL) require invasive parameter updates~\citep{shenfeld2025rlsrazoronlinereinforcement,huan2025doesmathreasoningimprove}, leading to issues such as catastrophic forgetting~\citep{li-etal-2024-revisiting,ding2025improvedsupervisedfinetuninglarge} or substantial computational overhead~\citep{liao2025enhancingefficiencyexplorationreinforcement}. Meanwhile, training-free prompting~\citep{wang-etal-2025-beyond-prompt} suffers from signal attenuation during forward propagation~\citep{wu2025lessunderstandingchainofthoughtlength}. Activation steering offers a promising alternative by directly modifying internal activations without changing model weights, but existing methods predominantly use a \emph{single} steering vector with \emph{fixed} intervention strength~\citep{venhoff2025understanding,ICLR2025_6e73c39c,jin-etal-2025-internal}. Such static, one-dimensional control limits their expressiveness and fails to capture the rich, multi-faceted structure of reasoning embedded in large models.


Decades of cognitive neuroscience research suggest a more flexible architecture: human cognition emerges from \textbf{modular} functional regions~\citep{1999The,2010Neural} coordinated by the \textbf{Prefrontal Cortex (PFC)} through dynamic routing of control signals~\citep{annurev:/content/journals/10.1146/annurev.neuro.24.1.167}. This perspective suggests a different blueprint, and highlights two key elements absent in current activation steering methods: a set of diverse reasoning primitives and a controller capable of selecting and composing them adaptively. Inspired by this, we ask: \textit{Can we build a PFC-like controller that adaptively awakens, routes, and composes latent cognitive capabilities through activation-level interventions during inference?}


Furthermore, recent advances in Representation Engineering~\citep{NEURIPS2024_fb3ad59a,postmus2024steering, alain2018understandingintermediatelayersusing} show that LLMs’ activation spaces contain interpretable, semantically meaningful directions to capture attributes, skills, and latent reasoning patterns~\citep{lee2025probingdifficultyperceptionmechanism,Marks2023TheGO,cyberey2025steering,Zhang_Wang_Li_Ao_He_2025,rimsky-etal-2024-steering}. These directions can be disentangled, composed, and manipulated to alter model behavior~\citep{fartale2025disentanglingrecallreasoningtransformer}. Such findings support the hypothesis that complex reasoning may be decomposable into multiple linear subspaces, each corresponding to a distinct cognitive capability. If true, then static single-vector steering is fundamentally misaligned with the structure of the model: what is needed is dynamic, compositional, and task-aware control.

Therefore, we propose a dynamic activation steering approach, termed \textbf{RISER} (\textbf{R}outer-based \textbf{I}ntervention for \textbf{S}teerable \textbf{E}nhancement of \textbf{R}easoning), which enables adaptive and compositional control of the model's internal reasoning process. RISER treats distinct activation patterns as reusable reasoning directions (which we call cognitive primitives). We first extract reasoning vectors that encode core cognitive functions to form a reusable capability library. Then, we introduce a lightweight Router module acting as a reasoning controller. Given an input query, the Router dynamically selects relevant vectors, determines optimal intervention strengths, and injects the combined representation into the model's forward pass (see Figure \ref{fig:concept}). We use RL to directly optimize the routing policy, effectively externalizing the model's implicit logic into explicit, interpretable control decisions.


    

Our experiments across seven diverse reasoning benchmarks demonstrate RISER’s effectiveness. It yields 3.4\%--6.5\% absolute zero-shot accuracy improvements while maintaining inference efficiency close to standard inference. Further analysis reveals that the RL-trained Router autonomously composes multiple primitives in meaningful ways, offering transparent insight into which cognitive capabilities are invoked and how they synergize for a given task. We summarize our primary contributions as follows:
\begin{itemize}
    \item We propose RISER, a plug-and-play activation intervention framework that keeps LLM parameters frozen while a lightweight Router dynamically selects and composes cognitive primitives, enabling precise and task-adaptive reasoning control.
    \item We develop a rigorous pipeline for eliciting high-quality reasoning vectors, incorporating LLM-Judge filtering and clustering to construct an orthogonal and disentangled library of cognitive primitives with verified steering effects.
    \item Extensive evaluation on seven benchmarks shows that we establish a new state-of-the-art for activation steering, closing the gap between inference-time control and heavy fine-tuning while offering interpretable insights into latent skill composition.
\end{itemize}

\section{Related Work}
\begin{figure*}[t]
    \centering
    \includegraphics[width=0.95\textwidth]{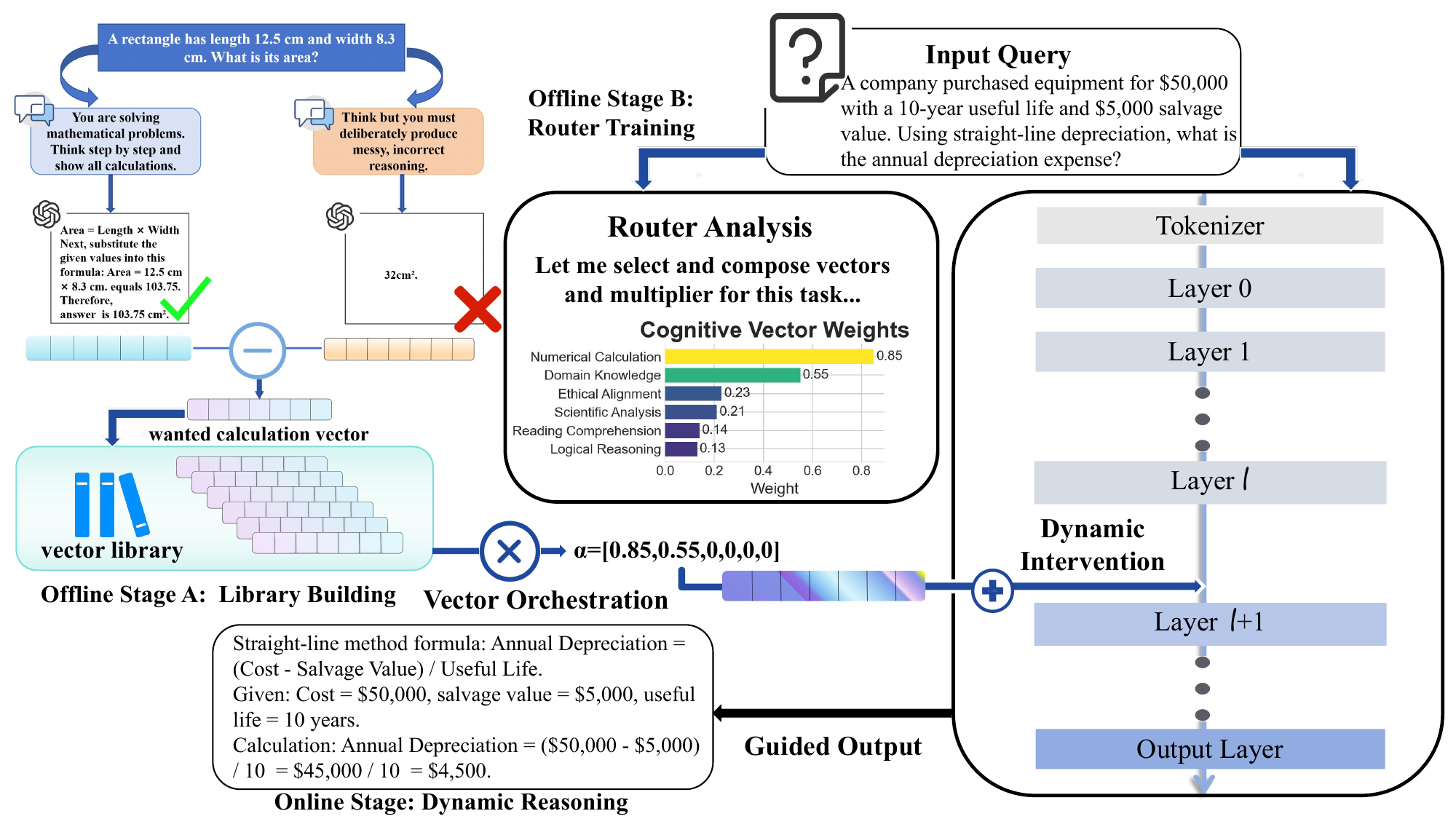}
    \caption{An overview of the RISER framework, illustrating the process of offline extraction of reasoning vectors and offline training of the Router, followed by online inference where the pre-trained Router dynamically selects and combines vectors  to intervene in the LLM's activation, guiding the final output.}
    \label{fig:framework}
\end{figure*}
\subsection{Activation Steering}
The study of linear representations has evolved from passive probing~\citep{alain2018understandingintermediatelayersusing,10.1162/coli_a_00422} to activation steering~\citep{2023Steering,2025Representation,DBLP:journals/corr/abs-2504-20020}, which enables active intervention on high-level concepts. Effective methods for concept vector extraction include Contrastive Activation Addition (CAA)~\citep{rimsky-etal-2024-steering} and SAEs~\citep{Cunningham2023SparseAF}. More recently, this paradigm has been extended to cognitive processes and reasoning, as it has been demonstrated that a reasoning vector extracted from one task can be applied to improve accuracy on another, thus confirming that reasoning capabilities are transferable~\citep{ICLR2025_6e73c39c,wang-etal-2025-beyond-prompt,venhoff2025basemodelsknowreason,zbeeb2025reasoningvectorstransferringchainofthought,valentino2025mitigatingcontenteffectsreasoning}. However, they typically apply a single, fixed vector with a manually-tuned strength for all inputs, failing to adapt to the specific demands of each task. While some work has introduced limited dynamics through gated activation~\citep{jin-etal-2025-internal} or strength calculation for single vectors~\citep{Zhang_Wang_Li_Ao_He_2025}, they do not address the challenge of composing multiple  capabilities with adaptive intervention strengths and rely on supervised objectives for training.

\subsection{Conditional Computation and Modular Networks}
Modular networks can be categorized into two main granularities: intra-model and inter-model. A prominent intra-model approach is the {Mixture-of-Experts (MoE)} architecture~\citep{article3,unknown1} that activates a small subset of experts for each input. At a coarser, inter-model granularity, researchers have explored task allocation among multiple independent LLMs~\citep{article2,Piskala_2024}. RISER applies the dynamic routing philosophy to a single, frozen LLM at the representation layer, offering greater flexibility and controllability.
\section{The RISER Framework}
\label{sec:framework}

\subsection{Overall Architecture}
\label{ssec:architecture}

As depicted in Figure~\ref{fig:framework}, RISER follows a simple offline-online split. Offline, we assume access to a compact library of reasoning vectors $\{\mathbf{v}_i\}_{i=1}^K$, each representing a reusable cognitive primitive. Given this library, we train a lightweight Router that learns to map internal states to compositions of these primitives (Section~\ref{ssec:optimization}). In Section~\ref{sec:elicitation}, we describe one concrete, data-driven instantiation process. In the online inference phase, the prepared components are used to intervene minimally but effectively in the LLM's forward computation. As an input query is processed up to a pre-determined intermediate layer $l$, we read the hidden state of the last token, $\mathbf{h}_l$, and feed it into the Router, which infers the immediate cognitive demands and outputs a composite reasoning vector by selecting and weighting a small subset of primitives. The resulting vector is injected back into the model to the activation at layer $l$, and maintained as a sustained cognitive priming during decoding.

\subsection{Router as a Dynamic Controller}
\label{ssec:Router}

The Router is a lightweight network that reads the model’s current hidden state and outputs an intervention in activation space. It receives the hidden state of the last token at the target layer $l$, denoted by $\mathbf{h}_l \in \mathbb{R}^{d}$, which serves as a natural proxy for the task's current reasoning demands. From this state, the Router produces two parallel outputs over the $K$ primitives at a sequence-level: a selection mask $\mathbf{w} \in \{0,1\}^K$ and a strength vector $\boldsymbol{\alpha} \in [0, \alpha_{\text{max}}]^K$. In practice, the selection head first outputs a probability vector $\mathbf{p} \in [0,1]^K$ via a Sigmoid activation, which is thresholded at inference time to obtain the binary mask $\mathbf{w}$ that specifies {which} capabilities to activate. During training, we apply the Gumbel-Sigmoid relaxation~\citep{DBLP:conf/iclr/JangGP17}. In parallel, the strength head predicts $\boldsymbol{\alpha}$, which specifies \emph{how strongly} to move along each selected direction. This dual-head design decouples the discrete choice of primitives from continuous intensity modulation, encouraging sparse yet flexible control. The final composite vector for injection is synthesized via a weighted summation over the primitive library:
\begin{equation}
    \mathbf{v}_{\text{inject}} = \sum_{i=1}^K w_i \cdot \alpha_i \cdot \mathbf{v}_i,
    \label{eq:v_inject}
\end{equation}
where $w_i$ and $\alpha_i$ are the $i$-th elements of $\mathbf{w}$ and $\boldsymbol{\alpha}$, respectively, and $\mathbf{v}_i$ is the $i$-th reasoning primitive. This vector is then injected into the LLM's forward pass via an element-wise addition to the last-token activation at layer $l$:
\begin{equation}
    \mathbf{h'}_l = \mathbf{h}_l + \mathbf{v}_{\text{inject}},
    \label{eq:intervention}
\end{equation}
where $\mathbf{h}_l$ is the original hidden activation and $\mathbf{h'}_l$ is the resulting steered activation. The model subsequently resumes its computation from layer $l{+}1$ onward using $\mathbf{h'}_l$, so that the remaining layers perform their usual processing under a slightly reoriented internal state toward the desired reasoning trajectory.

\subsection{Router Optimization}
\label{ssec:optimization}

Given the Router’s role, we optimize it directly from task-level feedback, enabling it to learn when and how to combine primitives to best serve downstream objectives.

\textbf{Supervised Warm-Up.} To avoid a cold-start regime, we first train the Router on a curated dataset derived from the vector library. For each training instance, we use grid search over the library to identify an intervention configuration $(\mathbf{w}^*, \boldsymbol{\alpha}^*)$ that successfully elicits the correct reasoning generation, establishing a robust baseline policy that knows which primitives tend to be useful in which contexts.

\textbf{Reinforcement Learning Refinement.} To move beyond the limitations of this static dataset and adapt to unseen task variations, we then fine-tune the Router with {Group Relative Policy Optimization (GRPO)~\citep{shao2024grpo}}. In this stage, the Router is free to explore the space of primitive compositions. We use an accuracy-based reward
\begin{equation}
r_i =
\begin{cases}
1, & \text{if the answer is correct},\\
0, & \text{otherwise},
\end{cases}
\label{eq:task-reward}
\end{equation}
which directly encourages policies that increase end-task accuracy. To ensure that interventions remain conservative and stable, we additionally impose a KL regularizer which at each decoding step $t$ computes
\begin{equation}
\small{
\mathcal{L}_{\mathrm{KL}} = \mathbb{E}_t \Big[ D_{\mathrm{KL}}\big(\pi_{\text{routed}}(\cdot \mid x, y_{<t}) \,\|\, \pi_{\text{base}}(\cdot \mid x, y_{<t})\big) \Big],}
\end{equation}
where $\pi_{\text{routed}}$ and $\pi_{\text{base}}$ denote the output distributions of the routed and base models. This regularizer discourages vector injections that induce large deviations from the base model’s behavior while still allowing beneficial changes, enabling the Router to learn primitive compositions without destabilizing the underlying model.


\section{Vector Elicitation Pipeline}
\label{sec:elicitation}

As effectiveness of the Router depends on the quality of the underlying primitives, we construct a concrete, data-driven vector elicitation pipeline. We first generate a broad candidate set of activation pairs following Contrastive Activation Additions. For a diverse range of reasoning tasks, we design paired prompts: a \textit{positive} prompt to elicit rigorous reasoning and a \textit{negative} prompt to suppress it. To mitigate noise, we integrate a quality-aware filtering mechanism using an LLM Judge (Claude-3.5-Sonnet~\citep{anthropic2024modelcardaddendum}) which evaluates the generated reasoning traces with strict inclusion criteria: only sample pairs where the positive generation achieves proficient reasoning and the negative one clearly lacks reasoning are retained.

\begin{figure}[t]
    \centering
    \includegraphics[width=1.0\linewidth]{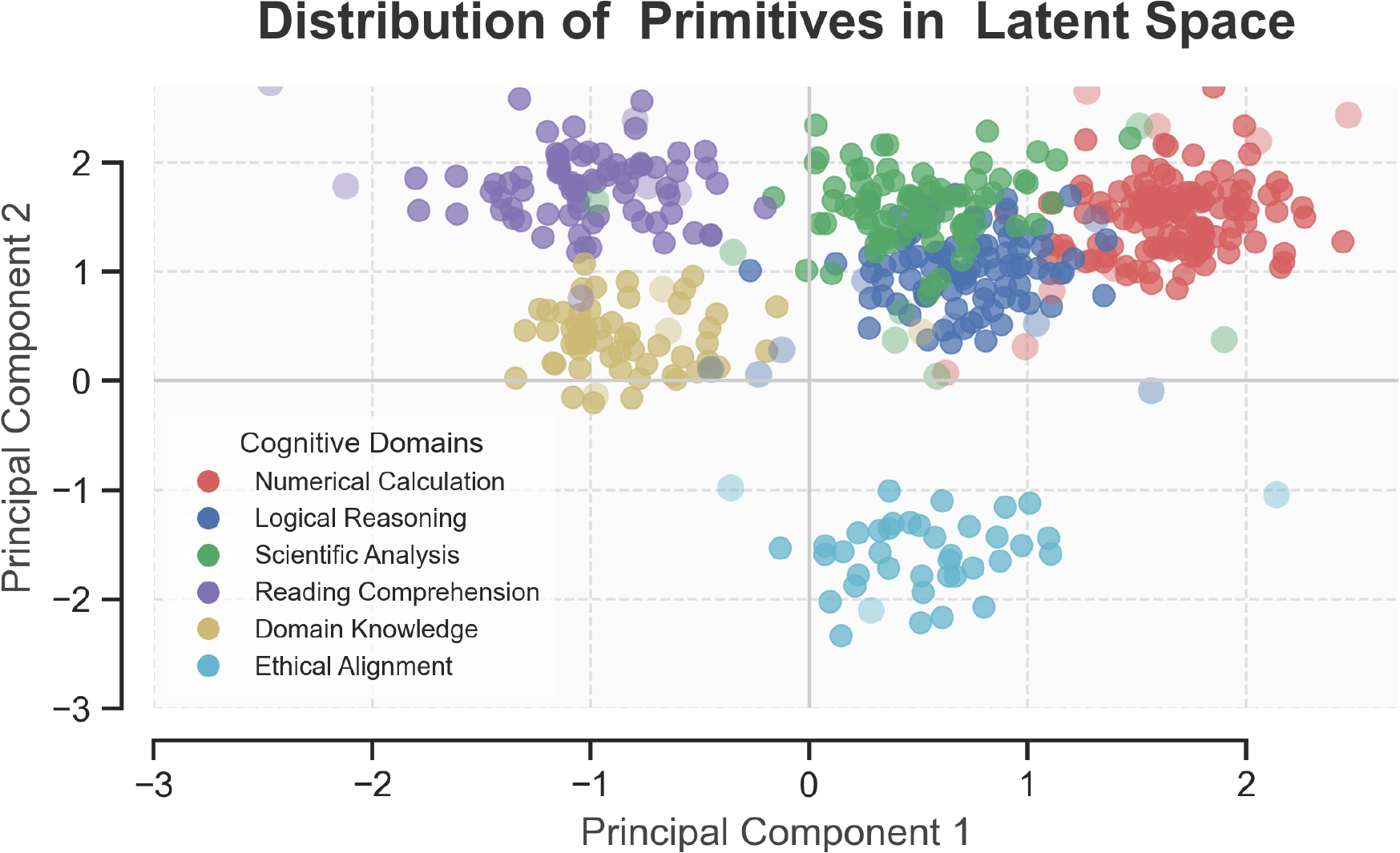}
    \caption{Latent space visualization of extracted vectors. We project the high-dimensional difference vectors onto a 2D plane using PCA. The visualization reveals naturally forming clusters, demonstrating that the extracted reasoning vectors possess strong semantic separability within the activation space.}
    \label{fig:pca}
\end{figure}

After filtering, we investigate the structure of the difference vectors ($\mathbf{h}^{+} - \mathbf{h}^{-}$) to build a compact set of representative reasoning directions. Our goal is to construct a small, interpretable, and practically controllable library suitable for downstream routing. To this end, we perform PCA visualization to examine the coarse geometry of the activation manifold. As shown in Figure~\ref{fig:pca}, the projections consistently exhibit several recurring high-density regions.

Rather than adopting a large number of clusters, we select major clusters that appear stably across random initializations and that jointly cover the dominant reasoning patterns observed in data (numerical reasoning, logical inference, ethical alignment, reading comprehension, scientific analysis, and domain-specific knowledge), maintaining a compact and human-understandable control space. Guided by the observation that the first six principal components account for over 85\% of the total variance in the extracted vectors, we set the cluster count $K=6$ and apply K-Means clustering to formally separate the difference vectors. Let $\mathcal{S}_i$ denote the set of sample indices assigned to the $i$-th cluster. The reasoning vector $\mathbf{v}_i \in \mathbb{R}^d$ for primitive $i$ is defined as the centroid:
\begin{equation}
\mathbf{v}_{i} = \frac{1}{|\mathcal{S}_i|} \sum_{j \in \mathcal{S}_i} (\mathbf{h}_{j}^{+} - \mathbf{h}_{j}^{-}),
\label{eq:vector_calculation}
\end{equation}
where $\mathbf{h}_{j}^{+}$ and $\mathbf{h}_{j}^{-}$ denote the positive and negative activations at layer $l$ for the $j$-th pair. We apply L2 normalization ($\mathbf{v}_{i} \leftarrow \mathbf{v}_{i}/\|\mathbf{v}_{i}\|_2$) to ensure consistent intervention magnitudes.

\begin{figure}[t]
    \centering
    \includegraphics[width=0.9\linewidth]{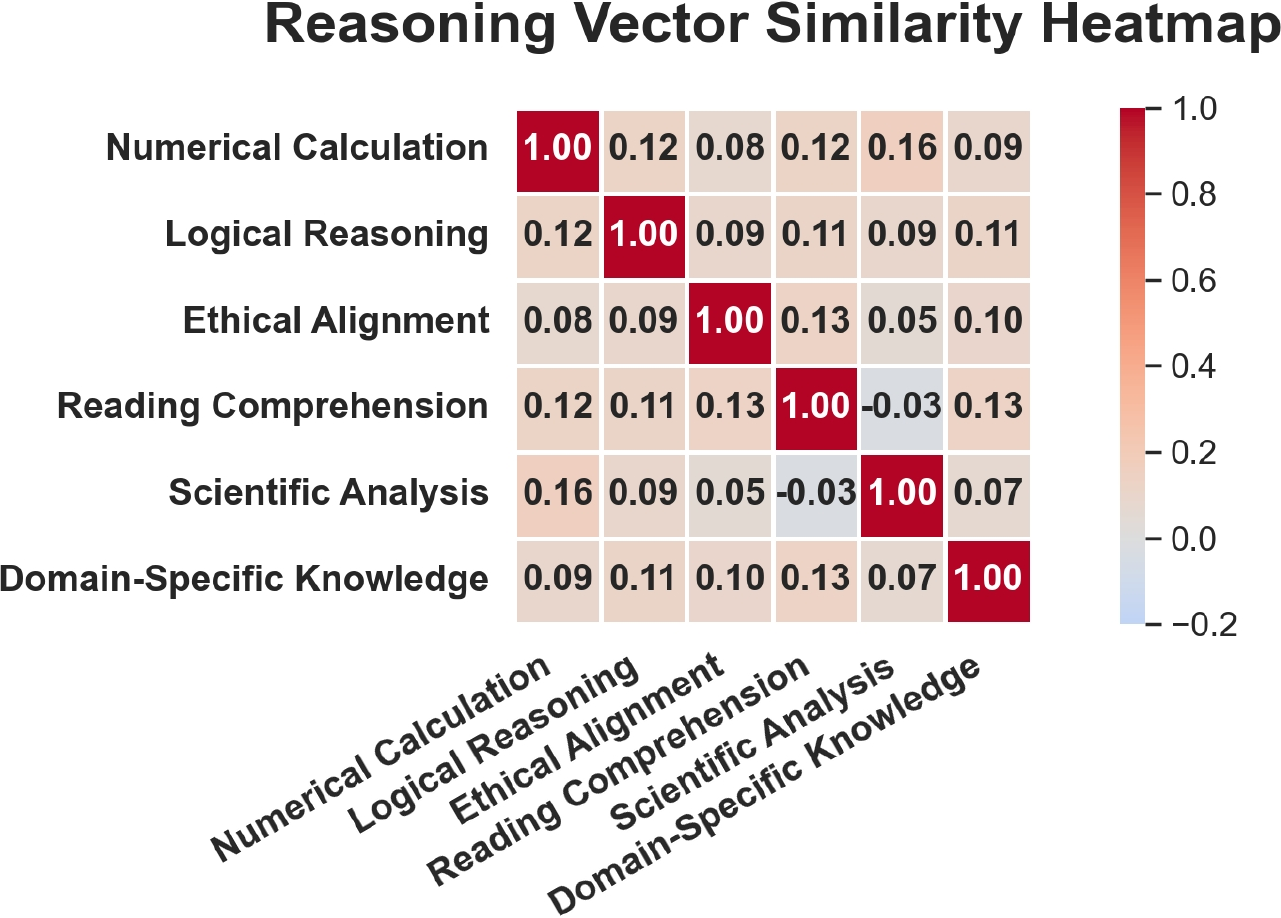}
    \caption{Reasoning vector library similarity heatmap. The low off-diagonal values confirm that the extracted vectors represent distinct and separable cognitive functions.}
    \label{fig:cosine}
\end{figure}

\begin{figure}[t]
    \centering
    \includegraphics[width=1\linewidth]{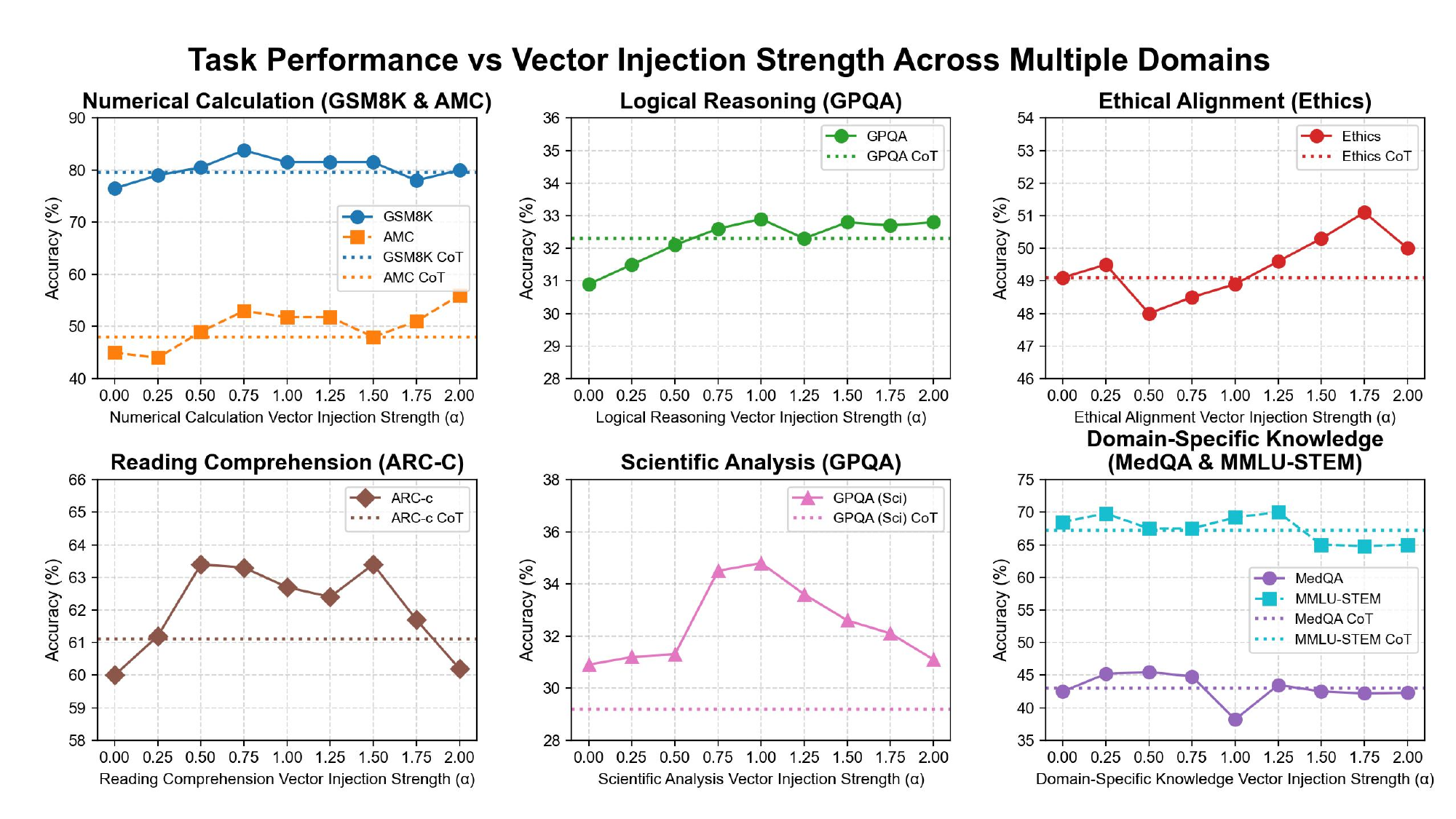}
    \caption{Static steering validation. The performance sensitivity to steering strength ($\alpha$) confirms that the extracted vectors effectively modulate specific reasoning behaviors.}
    \label{fig:singlevector}
\end{figure}

The resulting library shows that the vectors are nearly orthogonal, with an average pairwise cosine similarity $< 0.1$ (Figure~\ref{fig:cosine}), indicating that the selected directions represent distinct and minimally overlapping cognitive behaviors. Furthermore, as shown in Figure~\ref{fig:singlevector}, static steering experiments validate their functional efficacy: applying them can yield accuracy improvements on corresponding datasets. These findings confirm that our compact, engineering-driven extraction yields a set of effective and independently controllable reasoning modules.

\section{Experiments}
\subsection{Experimental Setup}
\label{ssec:setup}

We employ the Qwen2.5 family (7B-Instruct, 14B-Instruct, 32B-Instruct)~\citep{qwen2025qwen25technicalreport}  and Llama-3-8B-Instruct~\citep{grattafiori2024llama3herdmodels} as base models and, following prior work~\citep{chen2025personavectorsmonitoringcontrolling}, respectively select layers 20, 25, 40 and 13 as default layers for vector elicitation and intervention. We threshold gating selection probabilities at $0.7$, set the maximum intervention strength to $\alpha_{\max}=2.0$ and implement the Router as a lightweight, bottleneck-style Multi-Layer Perceptron (MLP) with approximately 5 million parameters ($<0.1$\% of base model), distinguishing our approach from simple linear classifiers while introducing negligible latency.
\begin{table*}[h]
\centering
\caption{Performance Comparison on Qwen2.5 Family and Llama-3 Models. We report accuracy (\%) and improvements ($\Delta$) over the baseline. \best{Best} and \second{Second} best results are labeled.}
\label{tab:performance}

\scriptsize
\setlength{\tabcolsep}{2pt}
\renewcommand{\arraystretch}{1.05}

\resizebox{\textwidth}{!}{%
\begin{tabular}{lcccccccc}
\toprule
\textbf{Dataset / Category} & \textbf{Base Model} & \textbf{CoT-Prompting} &
\textbf{Self-Consistency CoT} & \textbf{CAA} &\textbf{CAST} &\textbf{SAS} &
\textbf{FR-Ponder} & \textbf{Our Method (RISER)} \\
\midrule
\rowcolor{gray!15} \multicolumn{9}{l}{\textbf{Qwen2.5-7B-Instruct}} \\
MATH & 46.8 & 51.5 & \second{52.1} & 47.1 & 49.8 & 48.3 & 49.2 & \best{53.3(+6.5)} \\
GSM8K & 79.8 & \best{85.3} & \second{85.2} & 82.9 & 84.6 & 83.8 & 84.4 & \second{85.2(+5.4)} \\
\rowcolor{gray!10}
\textit{Average (Math/Logic)} & 63.3 & 68.4 & \second{68.7} & 65.0 & 67.2 & 66.1 & 66.8 & \best{69.3(+6.0)} \\
\addlinespace
GPQA & 31.0 & 31.2 & 33.2 & 31.9 & \second{33.7} & 32.6 & 33.0 & \best{36.8(+5.8)} \\
MMLU-Pro (in-dist.) & 44.1 & 44.0 & 44.2 & 46.2 & 47.5 & 46.9 & \second{47.7} & \best{50.3(+6.2)} \\
ARC-C & 63.7 & 63.3 & 64.4 & 63.3 & 65.3 & 64.1 & \second{65.8} & \best{67.2(+3.5)} \\
\rowcolor{gray!10}
\textit{Average (General)} & 46.3 & 46.2 & 47.3 & 47.1 & \second{48.8} & 47.9 & \second{48.8} & \best{51.4(+5.1)} \\
\addlinespace
Ethics & 48.6 & 49.3 & 48.7 & \best{53.2} & \second{52.4} & 51.0 & 50.2 & 52.1(+3.5) \\
TruthfulQA & 56.4 & 58.9 & 58.9 & \second{59.6} & 59.4 & 59.1 & 59.0 & \best{59.8(+3.4)} \\
\rowcolor{gray!10}
\textit{Average (Moral)} & 52.5 & 54.1 & 53.8 & \best{56.4} & 55.9 & 55.1 & 54.6 & \second{56.0(+3.5)} \\
\midrule
\rowcolor{gray!15} \multicolumn{9}{l}{\textbf{Qwen2.5-14B-Instruct}} \\
MATH & 55.6 & 58.9 & \second{60.4} & 56.4 & 59.0 & 57.4 & 58.2 & \best{61.8(+6.2)} \\
GSM8K & 86.5 & \second{90.5} & 89.7 & 88.6 & 90.1 & 89.2 & 90.0 & \best{90.8(+4.3)} \\
\rowcolor{gray!10}
\textit{Average (Math/Logic)} & 71.1 & 74.7 & \second{75.1} & 72.5 & 74.6 & 73.3 & 74.1 & \best{76.3(+5.2)} \\
\addlinespace
GPQA & 32.8 & 33.6 & 34.5 & 33.7 & \second{35.9} & 34.2 & 35.3 & \best{38.0(+5.2)} \\
MMLU-Pro (in-dist.) & 51.2 & 50.9 & 51.8 & 54.0 & 55.1 & 54.7 & \second{55.6} & \best{57.2(+6.0)} \\
ARC-C & 67.3 & 67.6 & 68.2 & 67.1 & 69.4 & 67.9 & \second{70.0} & \best{71.6(+4.3)} \\
\rowcolor{gray!10}
\textit{Average (General)} & 50.4 & 50.7 & 51.5 & 51.6 & 53.5 & 52.3 & \second{53.6} & \best{55.6(+5.2)} \\
\addlinespace
Ethics & 64.3 & 66.1 & 65.2 & \best{68.5} & 67.7 & 66.3 & 66.9 & \second{67.9(+3.6)} \\
TruthfulQA & 58.4 & 60.6 & 61.5 & 61.3 & \second{61.8} & 61.4 & 61.0 & \best{62.1(+3.7)} \\
\rowcolor{gray!10}
\textit{Average (Moral)} & 61.4 & 63.4 & 63.4 & \second{64.9} & 64.8 & 63.9 & 64.0 & \best{65.0(+3.6)} \\
\midrule
\rowcolor{gray!15} \multicolumn{9}{l}{\textbf{Qwen2.5-32B-Instruct}} \\
MATH & 57.7 & 60.8 & \second{61.9} & 58.8 & 60.9 & 59.6 & 60.4 & \best{63.2(+5.5)} \\
GSM8K & 90.9 & 93.2 & \second{93.5} & 91.9 & 93.0 & 92.6 & 92.7 & \best{93.9(+3.0)} \\
\rowcolor{gray!10}
\textit{Average (Math/Logic)} & 74.3 & 77.0 & \second{77.7} & 75.4 & 77.0 & 76.1 & 76.6 & \best{78.6(+4.3)} \\
\addlinespace
GPQA & 48.0 & 48.5 & 49.9 & 48.9 & \second{50.3} & 49.2 & 50.0 & \best{52.7(+4.7)} \\
MMLU-Pro (in-dist.) & 55.1 & 55.6 & 56.5 & 57.4 & 58.4 & 58.0 & \second{58.9} & \best{60.7(+5.6)} \\
ARC-C & 70.4 & 70.1 & 71.4 & 70.9 & 72.5 & 71.6 & \second{73.0} & \best{74.7(+4.3)} \\
\rowcolor{gray!10}
\textit{Average (General)} & 57.8 & 58.1 & 59.3 & 59.1 & 60.4 & 59.6 & \second{60.6} & \best{62.7(+4.9)} \\
\addlinespace
Ethics & 77.9 & 78.8 & 78.2 & \second{81.0} & \best{81.4} & 79.7 & 79.9 & \best{81.4(+3.5)} \\
TruthfulQA & 60.2 & 62.1 & 62.3 & 62.8 & \best{64.0} & 62.4 & 62.5 & \second{63.5(+3.3)} \\
\rowcolor{gray!10}
\textit{Average (Moral)} & 69.1 & 70.5 & 70.3 & 71.9 & \best{72.7} & 71.1 & 71.2 & \second{72.5(+3.4)} \\
\midrule

\rowcolor{gray!15} \multicolumn{9}{l}{\textbf{Llama-3-8B-Instruct}} \\
MATH & 30.9 & 33.6 & 34.1 & 32.8 & 33.2 & 32.4 & \second{34.2} & \best{35.4(+4.5)} \\
GSM8K & 84.5 & 88.2 & 87.3 & 88.5 & 88.7 & 88.0 & \second{88.9} & \best{89.1(+4.6)} \\
GPQA & 25.8 & 26.3 & \second{30.2} & 25.6 & 27.4 & 26.0 & 27.9 & \best{30.9(+5.1)} \\
TruthfulQA & 44.0 & 45.2 & 47.5 & 48.2 & \best{48.7} & 47.3 & 47.8 & \second{48.4(+4.4)} \\
\bottomrule
\end{tabular}
}
\end{table*}

	\textbf{Datasets.} The {Vector Elicitation} data consist of 500 problems randomly selected from MMLU~\citep{hendrycks2021measuringmassivemultitasklanguage}, each paired with positive and negative guiding prompts; During {Router Training}, the SFT phase uses an automated pipeline to extract and annotate 200 samples from MMLU, while the RL phase employs MMLU-Pro~\citep{10.5555/3737916.3740934} as the resource for reinforcement learning refinement. We split MMLU-Pro into 70\% training tasks for RL and 30\% held-out tasks for evaluation, with no question overlap. The {Evaluation Datasets} include benchmarks chosen to cover diverse reasoning types including math/logic reasoning (GSM8K~\citep{2021Training}, MATH~\citep{NEURIPS}), general reasoning (GPQA~\citep{Rein2023GPQAAG}, ARC-C~\citep{2018Think}, MMLU-Pro) and ethics and factual alignment (Ethics~\citep{hendrycks2021ethics},  TruthfulQA~\citep{lin-etal-2022-truthfulqa}).

\textbf{Baselines.} We compare against a set of baselines to quantify improvements: zero-shot base model; Chain-of-Thought (CoT) prompting~\citep{NEURIPS2022_9d560961}; Self-Consistency CoT~\citep{DBLP:conf/iclr/0002WSLCNCZ23} (with 5 samples and majority voting); CAA (static vector intervention with the best performance under different multipliers)~\citep{rimsky-etal-2024-steering}; CAST (conditional activation steering)~\cite{lee2025programmingrefusalconditionalactivation}; SAS (using sparse autoencoders for vector elicitation)~\cite{bayat2025steeringlargelanguagemodel} and FR-Ponder~\citep{ponder} (using a controller to regulate reasoning depth by selecting steering vectors).

\textbf{Ablation Settings.} To dissect component contributions, we evaluate several variants: \textbf{Direct GRPO Fine-tuning }(GRPO algorithm on the backbone model under an equivalent computational budget); \textbf{SFT-only Router} (Router trained only in the supervised phase without RL refinement); \textbf{Top-1 Vector Only} (select only the single highest-strength reasoning vector, disabling vector composition); and \textbf{Layer Sensitivity Analysis} (interventions applied at layers adjacent to the default layer as well as at earlier layer 5 and later layer 28 to assess sensitivity to intervention depth).

\textbf{Evaluation Metrics.} We report primary {task accuracy} and {token efficiency} measured by the total number of tokens generated.

\textbf{Implementation Details:}
For the SFT phase, we fine-tune the Router for 3 epochs with a learning rate of $5 \times 10^{-6}$. For RL phase we adopt a learning rate of $2 \times 10^{-6}$, a batch size of 128, a maximum context length of 8192 tokens during 2 epochs.
\subsection{Results}

{Table~\ref{tab:performance}} presents the comprehensive results on models, where RISER exhibits consistent performance gains across different model families. Focusing on the primary Qwen family, our method (RISER) achieves the highest average accuracy in the challenging \textbf{General Reasoning} category, significantly outperforming all other methods. In \textbf{Math/Logic Reasoning}, our method also outperforms the strong Self-Consistency CoT baseline. This demonstrates the framework's strong generalization and its ability to handle complex, multi-disciplinary tasks by dynamically composing capabilities. By learning to compose latent reasoning primitives only on one dataset, the Router acquires a transferable control strategy that generalizes across heterogeneous reasoning benchmarks.

\begin{table}[t]
\centering
\caption{Comprehensive ablation studies on key framework components and design choices. 
We report accuracy (\%) on representative datasets.}
\label{tab:ablation}
\resizebox{\columnwidth}{!}{%
\begin{tabular}{llccc}
\toprule
\textbf{Category} & \textbf{Model Variant / Setting} & \textbf{MATH} & \textbf{GPQA} & \textbf{TruthfulQA} \\
\midrule
\multicolumn{2}{l}{\textbf{Our Method (Full RISER @ L20)}} &{53.3} & {36.8} & {59.8} \\
\multicolumn{2}{l}{\textit{Direct GRPO (full-model RL fine-tuning)}} & 47.6 & 34.6 & 58.6 \\
\midrule
\parbox[t]{2cm}{\multirow{2}{*}{\textit{Training Ablation}}} &
\quad - w/o RL Refinement (SFT-only) & 49.4 & 31.2 & 54.6 \\
& \quad - w/o Composition (Top-1 Only) & 51.6 & 33.5 & 60.2 \\
\midrule
\parbox[t]{2cm}{\multirow{4}{*}{\textit{Layer Sensitivity}}} &
\quad - Early Layer (L5) & 48.5 & 31.5 & 55.0 \\
& \quad - Middle Layer (L19) & 52.1 & 35.5 & 59.5 \\
& \quad - Middle Layer (L21) & 51.8 & 34.6 & 59.6 \\
& \quad - Late Layer (L28) & 49.0 & 32.0 & 56.1 \\
\bottomrule
\end{tabular}}
\end{table}
We quantitatively analyze token efficiency on MATH and GPQA. Regarding efficiency, RISER requires only 1392 and 3056 tokens on MATH and GPQA, respectively, compared to 4033 and 6195 for CoT, realizing a 2–3× gain. While CoT generates reasoning-helpful external text, RISER mobilizes latent circuits for higher computational utilization, bypassing the need for verbose textual scaffolding to guide the trajectory.

We  compare our framework against static intervention CAA and other steering methods. The results clearly show the value of dynamic control. In the two categories requiring flexible, compositional reasoning (Math/Logic and General Reasoning), our dynamic Router significantly outperforms the static CAA baseline. Interestingly, in the {Moral Alignment} category, the static CAA or conditionally dynamic CAST baseline achieves the highest score, slightly edging out our method. This is likely because these tasks are highly uniform in their cognitive demands, and a strong, static application of the Ethical Alignment vector is highly effective. However, RISER still delivers substantial alignment improvements over all non-steering baselines. 

\begin{figure}
    \centering
    \includegraphics[width=1\linewidth]{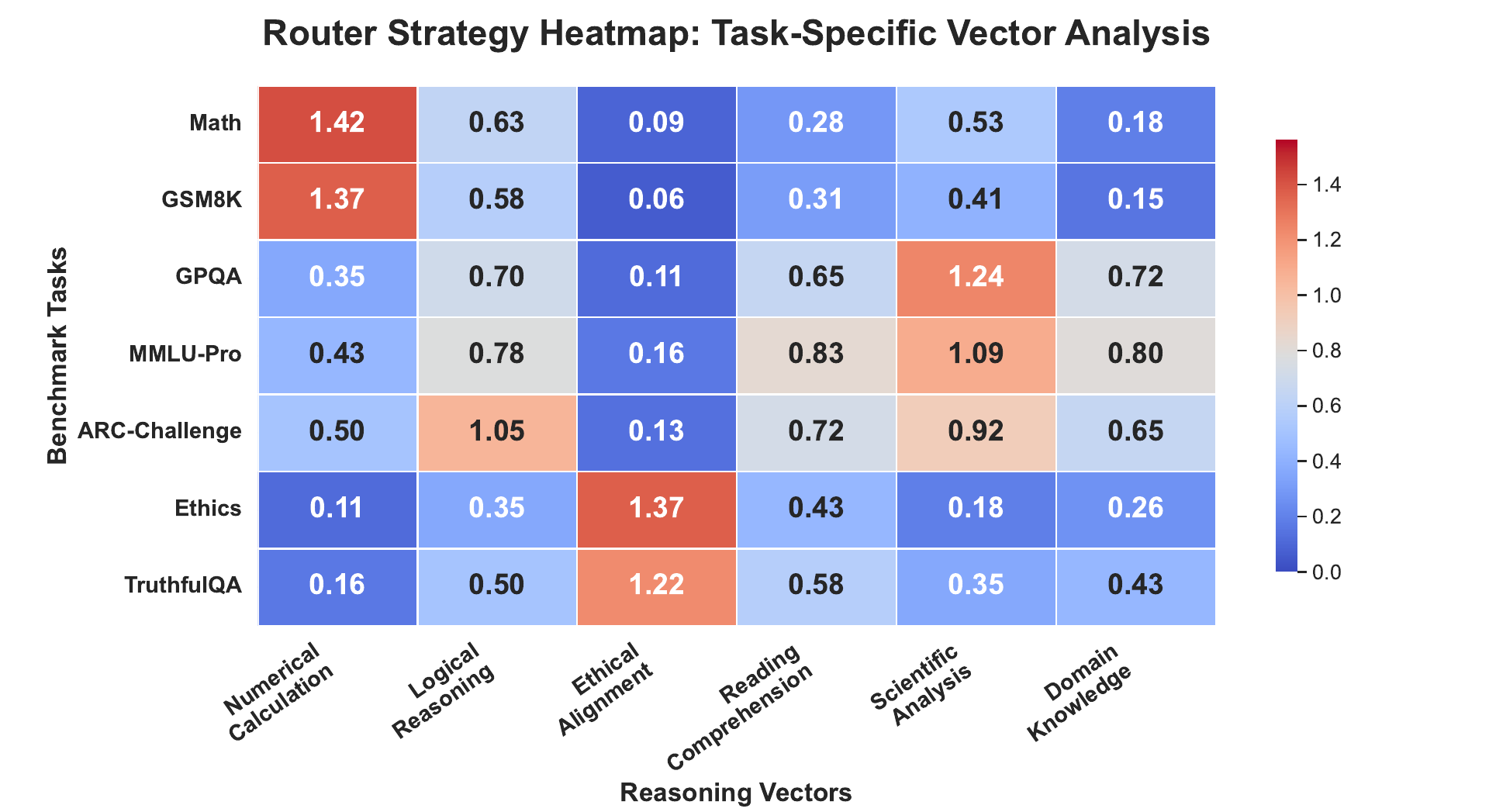}
    \caption{This heatmap shows the average strength assigned by the Router for each reasoning vector across different benchmarks and exhibits both logical specialization and complex composition.}
    \label{fig:heatmap}
\end{figure}

The Router strategy heatmap in Figure~\ref{fig:heatmap} also provides a cognitive map which intuitively demonstrates the explicit policy learned by the Router. On one hand, it learns a highly logical and specialized mapping: {MATH} and {GSM8K} tasks are strongly associated with the {Numerical Calculation} vector, while {Ethics} and {TruthfulQA} tasks correspond to the {Ethical Alignment} vector. On the other hand, when faced with complex cross-domain tasks (GPQA), the Router learns to autonomously compose multiple cognitive primitives. This provides direct evidence that the RL refinement phase externalized the LLM's implicit, synergistic strategies for complex problem-solving into an analyzable model.

\subsection{Ablation Studies}

We performed ablation studies on Qwen2.5-7B-Instruct (Table~\ref{tab:ablation}) to isolate the contributions of key components. 

\textbf{Comparison with Direct Fine-tuning:} A core question is whether the performance gains come from our RISER framework or simply from the RL training  itself. To answer this, we compare RISER against the {Direct GRPO} baseline. RISER  consistently outperforms the GRPO baseline in average accuracy across all three categories, which indicates that applying the same computational budget to train an external, dynamic reasoning controller is a more effective approach and validates its generalization advantage.

\textbf{Impact of RL Training:} The SFT-only Router significantly underperforms the full model, especially on complex benchmarks like GPQA and TruthfulQA, confirming that RL refinement is crucial for discovering synergistic vector compositions. 

\textbf{Necessity of Composition:} Restricting the Router to a single vector (Top-1 Vector Only) hurts performance on multi-disciplinary tasks, validating the critical role of vector orchestration. Conversely, on the homogeneous TruthfulQA, the Top-1 variant achieves a marginal gain, indicating that our framework correctly adapts to favor focused, single-vector interventions for monolithic tasks. 

\textbf{Layer Optimality:} Finally, Layer Sensitivity analysis identifies the middle layers as the optimal intervention site, showing robustness in adjacent layers but significant degradation at the model's input and output layers. This observation confirms the hypothesis that reasoning processes crystallize within the middle layers, acting as a critical bridge between the initial input processing in early layers and the final linguistic realization in later layers.

\subsection{Extensibility}
\label{sec:app_extensibility}

To investigate extensibility, we extend RISER to a different domain and introduce an additional primitive targeting code synthesis. Following the same vector elicitation and routing pipeline, we expand the Router’s output space to seven dimensions and perform a brief SFT phase on 200 examples, updating {only} the Router.  On HumanEval~\citep{chen2021evaluatinglargelanguagemodels}, the frozen base model achieves 56.3\% pass@1, while static CAA improves performance to 57.2\%. The extended Router over seven primitives further boosts accuracy to 59.9\% and does not significantly affect performance on the original reasoning benchmarks, indicating that newly added primitives can be integrated in a non-interfering manner. 

\subsection{Transferability Across Models}
\label{sec:transferability}

We further examine whether RISER can be reused beyond the backbone on which it is derived and evaluate cross-model transfer by directly applying a trained RISER configuration to a different target model. Within the same model family, transferring RISER across parameter scales remains effective, suggesting that both the learned vector library and the Router’s composition strategy align reasonably well across scales. In contrast, transferring across different model families provides no benefit, indicating that the primitive directions and routing policy are tightly coupled to model-specific representation geometry and activation statistics. These results indicate that transfer is promising when the underlying activation manifolds are sufficiently aligned, but not across heterogeneous architectures. Full analysis are in Appendix~\ref{sec:appendix_transferability2}.
\begin{table}[h]
    \centering
    \caption{{Cross-Model Transferability on MATH.} Off-diagonal entries show transfer results.}
    \label{tab:transferability_math_realistic}
    \resizebox{\linewidth}{!}{
    \begin{tabular}{l|cccc}
        \toprule
        & \multicolumn{4}{c}{\textbf{Source Router (Trained on)}} \\
        \cmidrule(lr){2-5}
        \textbf{Target Model (Inference)} & \textbf{Qwen-7B} & \textbf{Qwen-14B} & \textbf{Qwen-32B} & \textbf{Llama-3-8B} \\
        \midrule
        \textbf{Qwen2.5-7B} & \textbf{53.3} & 51.7 \small{(+4.9)} & 52.1 \small{(+5.3)} & 46.5 \small{(-0.3)} \\
        \textbf{Qwen2.5-14B} & 58.1 \small{(+2.5)} & \textbf{61.8} & 60.4 \small{(+4.8)} & 55.8 \small{(+0.2)} \\
        \textbf{Qwen2.5-32B} & 59.8 \small{(+2.1)} & 60.6 \small{(+2.9)} & \textbf{63.2} & 57.5 \small{(-0.2)} \\
        \midrule
        \textbf{Llama-3-8B} & 30.5 \small{(-0.4)} & 31.1 \small{(+0.2)} & 30.8 \small{(-0.1)} & \textbf{35.4} \\
        \bottomrule
    \end{tabular}
    }
\end{table}

\section{Conclusion}
RISER demonstrates that LLM reasoning can be effectively enhanced by orchestrating latent activations, offering a computationally efficient alternative to weight modification or verbose prompting. By learning explicit, RL-optimized policy, our framework achieves significant performance gains while validating the existence of steerable cognitive primitives within frozen models. This approach shifts the focus from surface-level text generation to internal state management, establishing a viable path toward more controllable and resource-efficient AI systems.

\section{Limitations}

Our framework, while effective, is constrained by its reliance on reactivating latent capabilities, making its performance bounded by the quality of the base model’s pre-training. The construction of a capability library with a fixed number of clusters further reflects an engineering trade-off: it stabilizes control but may reduce semantic granularity, potentially oversimplifying the underlying activation manifold for highly nuanced tasks. Moreover, the extracted vectors predominantly capture broad domain-level reasoning patterns due to the natural clustering structure of the model’s activation space. Future work can focus on disentangling these into finer-grained, domain-agnostic atomic skills, automating the discovery of such primitives, and exploring hierarchical routing mechanisms to achieve more precise control over complex reasoning chains. 
Finally, as RISER operates by modifying internal activations, careless application without proper constraints could lead to unintended behavioral shifts. In this work, we restrict our analysis to controlled benchmark settings, and future deployment-oriented use would require additional safety and alignment evaluation.

\bibliography{custom}

\appendix

\section{Case Study}

\begin{figure}
    \centering
    \includegraphics[width=1.02\linewidth]{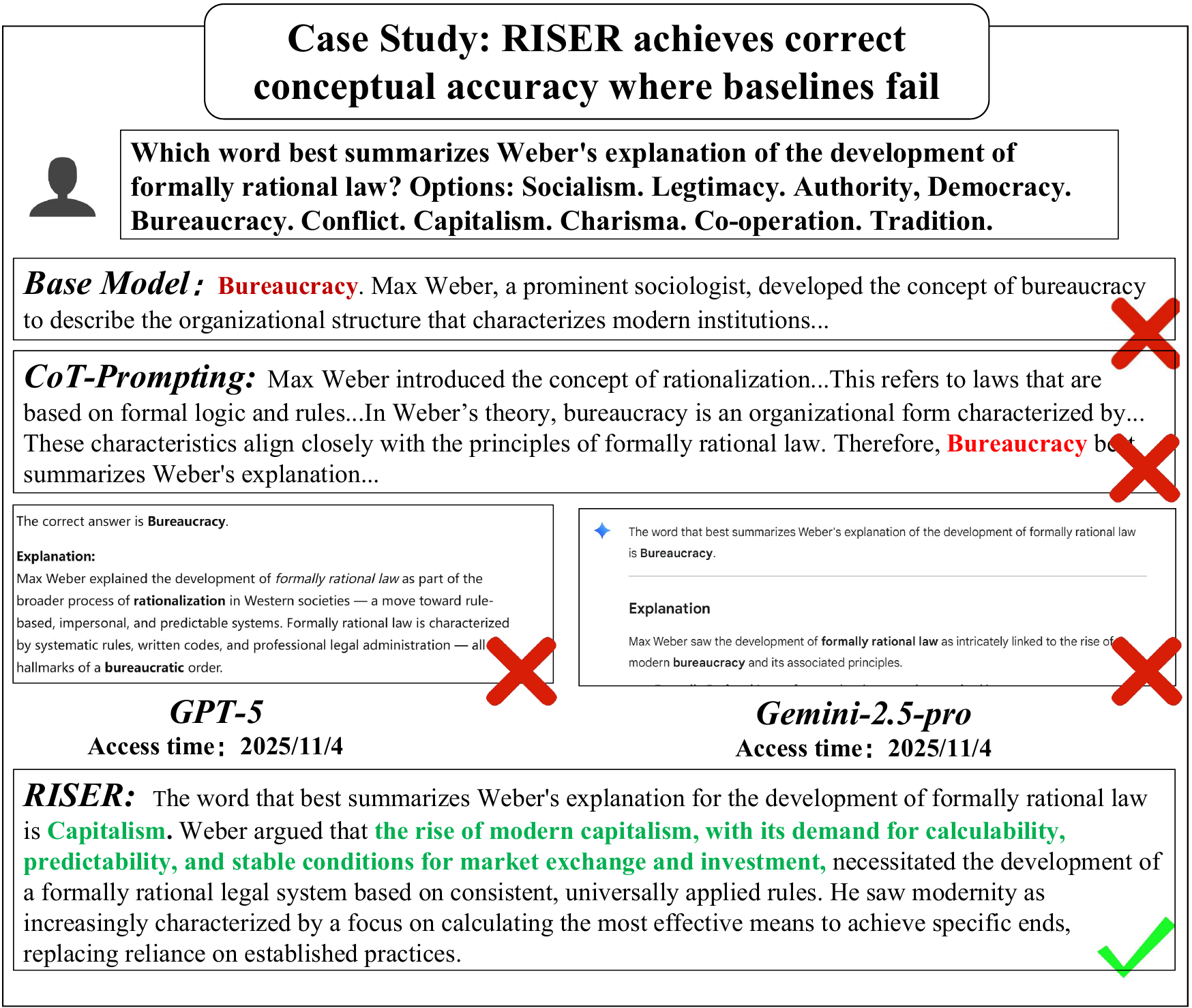}
    \caption{All baselines choose incorrectly, while RISER selects the correct answer and grounds its explanation in the rise of modern capitalism.}
    \label{fig:case}
\end{figure}

Figure~\ref{fig:case} presents a qualitative comparison where RISER successfully navigates conceptual traps. While the base model, CoT prompting, and even stronger frontier models are lured into selecting Bureaucracy by misleading lexical cues despite generating superficially coherent rationales, RISER correctly identifies that Capitalism aligns with Weber's theory. Analysis of the routing weights reveals that RISER selectively amplifies logical reasoning and domain knowledge primitives while suppressing irrelevant directions, effectively steering the model's focus. 

\section{Transferability Across Domains}
\label{sec:app_transferability1}
We further evaluated RISER directly on the Big-Bench Hard (BBH) benchmark~\citep{suzgun2022challengingbigbenchtaskschainofthought} without any additional fine-tuning. Despite the significant distribution shift moving from knowledge-centric exams to pure symbolic tasks, RISER outperformed the base model by 2.8\%. Crucially, the Router autonomously adapted its strategy by prioritizing general-purpose primitives, specifically Logical Reasoning and Reading Comprehension. This transferability suggests that these primitives capture content-agnostic cognitive mechanisms akin to fluid intelligence and the system successfully identifies that the underlying computational demand remains constant even when the surface-level domain changes, verifying that RISER orchestrates genuine, transferable mental skills.

\section{Transferability Across Models}
\label{sec:appendix_transferability2}

To delineate the generalization boundaries of the RISER framework, we systematically investigate the transferability of learned Routers across distinct model families and varying parameter scales in Table~\ref{tab:transferability_math_realistic}. Specifically, we apply a Router trained on a source model directly to a target model without any additional tuning. This experiment aims to determine whether the learned cognitive compositions capture universal reasoning patterns or remain specific to the internal representation space of the source architecture.

\subsection{Cross-Architecture Transferability}
We first evaluate Router transferability between distinct model families, specifically exchanging Routers between Qwen2.5-7B and Llama-3-8B. Empirical results demonstrate negligible generalization, with performance regressing to near-random or baseline levels.

This failure suggests that the learned reasoning vectors and routing policies are inextricably coupled to the specific manifold of each model family. We attribute this incompatibility to three primary factors. First, stochastic pre-training induces {manifold misalignment}, where semantically similar concepts acquire arbitrary geometric orientations in high-dimensional space, precluding natural isometry between families without explicit alignment. Second, variations in pre-training corpora and tokenizers yield {divergent activation statistics}, causing source vectors to map onto low-density or undefined regions in the target manifold. Third, {architectural inductive biases}---arising from structural differences such as Grouped-Query Attention versus Multi-Head Attention---fundamentally reshape the activation landscape geometry, rendering direct vector transplantation mathematically invalid.

\subsection{Intra-Family Transferability: Scale Invariance}
Conversely, transferring Routers within the same model family (e.g., across the Qwen2.5 series) yields substantial efficacy, indicating a shared semantic alignment across scales. We observe two distinct phenomena based on the direction of transfer.

\vspace{0.5em}
\noindent\textbf{Large-to-Small Transfer (Inference-Time Distillation).}
Transferring a Router from a larger model (e.g., 32B) to a smaller one (e.g., 7B) results in significant accuracy gains of up to \textbf{+5\%}. We posit that the Router derived from the larger model encapsulates more precise and robust cognitive strategies. Deploying this advanced policy on a smaller model functions as a form of inference-time distillation, effectively guiding the smaller model to navigate complex reasoning pathways that it fails to autonomously discover due to limited capacity.

\vspace{0.5em}
\noindent\textbf{Small-to-Large Transfer (Feature Consistency).}
Transferring from smaller to larger models also confers meaningful improvements, typically enhancing accuracy by \textbf{+2--3\%}. This finding demonstrates scalable feature consistency, implying that the fundamental cognitive directions identified in smaller parameter regimes remain preserved and refined in larger models. Consequently, the lightweight Router maintains its steering effectiveness even as the backbone capacity increases, highlighting the hierarchical stability of the learned representations within the same lineage.

\section{Implementation \& Training Details}
\label{sec:details}
\subsection{Hyperparameters}
\begin{table}[h]
    \centering
    \caption{Hyperparameters and implementation details for RISER training.}
    \label{tab:hyperparameters}
    \resizebox{0.95\linewidth}{!}{
    \begin{tabular}{lc}
        \toprule
        \textbf{Hyperparameter} & \textbf{Value} \\
        \midrule
        \multicolumn{2}{l}{\textit{Model Architecture \& Optimization Strategy}} \\
        Advantage Estimator & GRPO \\
        Base LLM Trainable & False \\
        Router Trainable & True \\
        
        \midrule
        \multicolumn{2}{l}{\textit{Training Configuration}} \\
        SFT Learning Rate & $5 \times 10^{-6}$ \\
        SFT Epochs & 3 \\
        RL Learning Rate & $2 \times 10^{-6}$ \\
        Total RL Epochs & 2 \\
        Global Batch Size & 128 \\
        Max Context Length & 8192 \\
        Data Shuffling & True \\
        
        \midrule
        \multicolumn{2}{l}{\textit{Reward \& KL Divergence}} \\
        Reward Type & Accuracy (0/1) \\
        KL Loss Coefficient & 0.001 \\
        KL Loss Type & Low Variance KL \\
        
        \midrule
        \multicolumn{2}{l}{\textit{Rollout \& Generation}} \\
        Temperature & 1.5 \\
        Top-k & -1 (Disabled) \\
        Do Sample & True \\
        Number of Rollouts ($N$) & 8 \\
        Max Batched Tokens & 8192 \\
        
        \midrule
        \multicolumn{2}{l}{\textit{Infrastructure \& Parallelism}} \\
        Tensor Parallel Size & 4 \\
        GPUs per Node & 4 \\
        Number of Nodes & 1 \\
        \bottomrule
    \end{tabular}
    }
\end{table}

In Table~\ref{tab:hyperparameters}, we list the training configuration and used hyperparameters during our experiment.  Full framework code will be released once accepted. Experiment is conducted on RTX5090 GPU.
We report accuracy for each benchmark and average performance across task groups and results are computed from random seeds (average performance on 3 runs) and report absolute accuracy and relative improvements over baselines.
\subsection{Datasets Configuration}
\label{ssec:datasets_config}

Our training pipeline consists of three distinct phases, each utilizing a specific data strategy to ensure the robustness and capabilities of the RISER framework.

\paragraph{Phase 1: Reasoning Vector Elicitation Data.}
To construct a comprehensive library of cognitive primitives, we require a dataset that covers a broad spectrum of reasoning types. We constructed the elicitation dataset by performing \textbf{random sampling} of 500 examples from the \textbf{MMLU} benchmark~\citep{hendrycks2021measuringmassivemultitasklanguage}. Given MMLU's inherent breadth across diverse subjects, ranging from elementary mathematics to professional law, this random sampling strategy ensures that the extracted vectors cover a diverse spectrum of cognitive reasoning patterns without introducing domain-specific bias. 

For each question, we generated several paired activation states using the \textit{Positive} and \textit{Negative} prompts (see Appendix~\ref{sec:appendix_prompts}). To ensure quality, we applied the LLM-Judge filtering mechanism described in Section~\ref{sec:elicitation}, retaining only pairs that exhibit a significant gap in reasoning rigor, paving the path for further experiments.

\paragraph{Phase 2: Router SFT Data (Oracle Label Synthesis).}
For the supervised warm-up, we utilized a separate set of 200 MMLU samples (non-overlapping with the elicitation set). To synthesize the ground-truth "Oracle Labels" for training the Router, we employed a \textbf{constrained grid search} mechanism. 
For each query, we first ranked the primitives based on their individual efficacy and pre-selected the \textbf{top-2} candidates. Within this reduced subspace, we performed a fine-grained grid search over the intervention strength $\alpha$, discretizing the value with a step size of \textbf{0.1} (ranging from 0 to $\alpha_{\max}$). The configuration $(\mathbf{w}^*, \boldsymbol{\alpha}^*)$ that elicited the correct response with the highest confidence was selected as the supervisory target. This approach efficiently provides high-quality initialization for the Router without the computational cost of an exhaustive search over the entire combinatorial space.

\paragraph{Phase 3: RL Refinement Data.}
To further optimize the Router for complex composition and generalization, we employed the \textbf{MMLU-Pro} benchmark. MMLU-Pro presents a significantly harder challenge with distractor options and complex reasoning chains, providing a steeper gradient for reinforcement learning compared to standard MMLU. We randomly split the dataset into a Training Set (70\%) for the GRPO algorithm and a Held-out Set (30\%) for validation. Importantly, we strictly ensured \textbf{no question overlap} between the RL training set and the final evaluation benchmarks (Table~\ref{tab:performance}) to prevent data leakage and ensure fair evaluation.

\section{RISER Inference Procedure}
Algorithm~\ref{alg:riser_inference} provides pseudocode for our inference pipeline: we invoke the Router once after the prompt prefill to compute an injection vector $v_{\text{inject}}$, and then reuse the same $v_{\text{inject}}$ as an additive intervention at layer $l$ for the last-token activation at every decoding step.

\label{app:riser_inference}

\begin{algorithm}[t]
\caption{RISER inference}
\label{alg:riser_inference}
\begin{algorithmic}[1]
\Require Frozen LLM $f_\theta$ ($L$ layers), Router $g_\phi$, primitives $V=\{v_i\}_{i=1}^K$, layer index $l$,
threshold $\tau$, $\alpha_{\max}$, prompt tokens $x$, max steps $T$
\Ensure Generated tokens $y$

\State \textbf{Prefill (compute $v_{\text{inject}}$ once)}:
\State $h_l \gets \textsc{ForwardToLayer}(f_\theta, x, l)$ \Comment{last-token state at layer $l$}
\State $(p,\alpha) \gets g_\phi(h_l)$ \Comment{$p\in[0,1]^K,\ \alpha\in\mathbb{R}^K$}
\For{$i=1$ to $K$}
  \State $w_i \gets \mathbb{I}[p_i>\tau]$
  \State $\alpha_i \gets \mathrm{clip}(\alpha_i, 0, \alpha_{\max})$
\EndFor
\State $v_{\text{inject}} \gets \sum_{i=1}^K w_i\,\alpha_i\,v_i$

\State \textbf{Decoding (reuse the same $v_{\text{inject}}$ at every step)}:
\State $y \gets [\ ]$
\For{$t=1$ to $T$}
  \State $h_l^{(t)} \gets \textsc{ForwardToLayer}(f_\theta, x \Vert y, l)$
  \State $\tilde{h}_l^{(t)} \gets h_l^{(t)} + v_{\text{inject}}$
  \State $\mathrm{logits}^{(t)} \gets \textsc{ContinueFromLayer}(f_\theta, \tilde{h}_l^{(t)}, l)$
  \State $y_t \gets \textsc{DecodeToken}(\mathrm{logits}^{(t)})$
  \State $y \gets y \Vert [y_t]$
  \If{$y_t$ is EOS}
    \State \textbf{break}
  \EndIf
\EndFor
\State \Return $y$
\end{algorithmic}
\end{algorithm}

\section{Sensitivity Analysis of Cluster Count ($K$)}
\label{app:sensitivity_k}

We fixed the size of the cognitive primitive library at $K=6$ based on the observation that the first six principal components account for over 85\% of the variance in the extracted difference vectors. To empirically validate this choice and assess the sensitivity of RISER to the granularity of the primitive library, we conducted an ablation study with varying cluster counts $K \in \{4, 6, 8, 12\}$. We evaluated these variants on Qwen2.5-7B-Instruct using three benchmarks that require distinct reasoning capabilities: GSM8K (Math), GPQA (General/Scientific), and TruthfulQA (Safety/Alignment). All other hyperparameters were held constant.
\begin{table}[h]
    \centering
    \resizebox{1\linewidth}{!}{
        \begin{tabular}{lccccc}
        \toprule
        \textbf{Primitives ($K$)} & \textbf{Variance Explanation (PCA)} & \textbf{GSM8K} & \textbf{GPQA} & \textbf{TruthfulQA} & \textbf{Avg.} \\
        \midrule
        $K=4$ & 72.4\% & 83.1 & 33.5 & 56.2 & 57.6 \\
        \textbf{$K=6$ (Ours)} & \textbf{86.1\%} & \textbf{85.2} & \textbf{36.8} & 59.8 & \textbf{60.6} \\
        $K=8$ & 89.3\% & 85.0 & 36.4 & \textbf{60.1} & 60.5 \\
        $K=12$ & 93.5\% & 84.6 & 35.9 & 59.5 & 60.0 \\
        \bottomrule
        \end{tabular}
    }
    \caption{\textbf{Sensitivity Analysis of Cluster Count ($K$).} We report the percentage of variance explained by the top-$K$ principal components and the zero-shot accuracy across three representative benchmarks. $K=6$ strikes the optimal balance between performance and model complexity.}
    \label{tab:sensitivity_k}
\end{table}

The results, summarized in Table~\ref{tab:sensitivity_k}, demonstrate that $K=6$ is not merely an arbitrary choice but a local optimum for performance. 

\textbf{Under-clustering ($K=4$):} Reducing the number of primitives leads to a noticeable performance drop ($-3.0\%$ average accuracy compared to $K=6$). With only 72.4\% of the variance explained, distinct cognitive functions (e.g., \textit{Numerical Calculation} and \textit{Logical Reasoning}) are forced to merge into coarser centroids. This \textit{semantic collision} reduces the precision of the steering vectors, preventing the Router from isolating the specific capability required for specialized tasks like GSM8K.

\textbf{Over-clustering ($K=8, 12$):} Increasing $K$ beyond 6 yields diminishing returns. While the explained variance increases to 93.5\% at $K=12$, the downstream accuracy plateaus or slightly degrades. We attribute this to two factors: 

(1) \textit{Redundancy}: Higher $K$ values introduce collinear vectors that represent fine-grained nuances rather than distinct skills, reducing the orthogonality of the library. 

(2) \textit{Optimization Difficulty}: A larger action space complicates the RL exploration process. The Router struggles to distinguish between redundant vectors given the sparse reward signal, leading to less stable policies. 

Consequently, $K=6$ provides the most robust trade-off, ensuring sufficient semantic coverage to handle diverse tasks while maintaining a compact and orthogonal action space for efficient Router learning.
\section{Prompt for Reasoning Quality Isolation}
\label{sec:appendix_prompts}

To extract reasoning vectors that encode {cognitive rigor} rather than mere verbosity, we employ a Reasoning Fidelity Contrast strategy. The goal is to isolate the difference between \textbf{verified execution} and \textbf{plausible generation}, ensuring the extracted vector promotes efficiency by substituting internal computation for external token generation.

\subsection{Vector Elicitation Prompts}
\label{sec:elicitation_prompts}

\definecolor{poscolor}{RGB}{235, 245, 235} 
\definecolor{negcolor}{RGB}{245, 235, 235} 

\begin{promptbox}[colback=poscolor]{Positive}
\textbf{Role:} You are a meticulous logician focused on absolute precision.\\
\textbf{Task:} Derive the answer to the following question using proof-level rigor.

\textbf{Instructions:}
\begin{itemize}
    \setlength\itemsep{0em}
    \item \textbf{Derive:} Do not just state facts; deduce them from axioms or given data.
    \item \textbf{Verify:} Check each intermediate calculation or logic jump for errors.
    \item \textbf{Precision:} Prioritize correctness over fluency. Reject any heuristic shortcuts.
    \item \textbf{Output:} Provide a  sound explanation.
\end{itemize}

\textbf{Question:}\\
\{\{QUESTION\}\}

\textbf{Rigorous Derivation:}
\end{promptbox}

\vspace{0.8em}

\begin{promptbox}[colback=negcolor]{Negative}
\textbf{Role:} You are a fluent conversationalist acting on "autopilot".\\
\textbf{Task:} Provide a plausibly sounding answer based on surface-level associations.

\textbf{Instructions:}
\begin{itemize}
    \setlength\itemsep{0em}
    \item \textbf{Flow:} Write whatever comes to mind first based on language patterns.
    \item \textbf{Approximate:} Do not perform actual calculations or verification. Use "ballpark" figures.
    \item \textbf{Plausibility:} The answer should \textit{sound} correct to a layperson, even if the logic is flawed.
    \item \textbf{Output:} Generate a coherent but unverified response (simulate a hallucination if necessary).
\end{itemize}

\textbf{Question:}\\
\{\{QUESTION\}\}

\textbf{Plausible Response:}
\end{promptbox}
\subsection{LLM Judge Filtering Criteria}
\label{sec:judge_criteria}

We employ an LLM Judge to ensure the vector subtraction captures the {quality gap}. The judge filters for pairs where the positive response is logically sound (Score $>80$) and the negative response lacks actual reasoning depth (Score $<20$), while maintaining structural similarity.

\begin{promptbox}{Evaluation Prompt}
You are evaluating the \textbf{cognitive integrity and reasoning quality} of two AI responses.
\textbf{Question:} \{\{QUESTION\}\}

\textbf{Response A:} \{\{RESPONSE\_A\}\}
\textbf{Response B:} \{\{RESPONSE\_B\}\}

\textbf{Criteria:}
\begin{enumerate}
    \setlength\itemsep{0em}
    \item \textbf{Soundness (A):} Does Response A contain a valid, verified logical path to the correct answer? (Score 0-100). \textit{High score means rigorous logic.}
    \item \textbf{Logical Validity (B):} Does Response B demonstrate actual step-by-step derivation? (Score 0-100). \textit{Low score means it relies on guessing, hallucinations, or surface-level associations without real logic.}
    \item \textbf{Structural Parity:} Are both responses roughly similar in length? (Pass/Fail).
\end{enumerate}

\textbf{Output JSON:}
\begin{verbatim}
{
 "score_A_soundness": <int>,
 "score_B_validity": <int>,
 "structural_parity": <bool>,
 "valid_pair": <bool> 
 // True ONLY if:
 // score_A_soundness > 80 AND 
 // score_B_validity < 20 AND 
 // structural_parity is True.
}
\end{verbatim}
\end{promptbox}

\end{document}